\documentclass[10pt,twocolumn,letterpaper]{article}
\usepackage[margin=0.75in,columnsep=0.25in]{geometry}
\usepackage[T1]{fontenc}
\usepackage[utf8]{inputenc}
\usepackage{lmodern,microtype}
\usepackage{amsmath,amssymb,booktabs,array,graphicx,tabularx}
\usepackage{algorithm,algpseudocode}
\usepackage{tikz}
\usetikzlibrary{arrows.meta,positioning}
\usepackage[font=small,labelfont=it]{caption}
\usepackage{xurl}
\usepackage{titlesec}
\titleformat{\section}{\normalfont\fontsize{12}{14}\selectfont\bfseries\raggedright}{\thesection}{0.65em}{}
\titleformat{\subsection}{\normalfont\normalsize\bfseries\raggedright}{\thesubsection}{0.65em}{}
\titlespacing*{\section}{0pt}{2.2ex plus .5ex minus .2ex}{1ex plus .2ex}
\titlespacing*{\subsection}{0pt}{1.7ex plus .4ex minus .2ex}{.7ex plus .2ex}
\makeatletter
\renewcommand{\@maketitle}{%
  \newpage
  \begin{center}%
    \hrule height 2pt
    \vskip 10pt
    {\fontsize{20}{24}\selectfont\bfseries \@title\par}%
    \vskip 10pt
    \hrule height .5pt
    \vskip 12pt
    {\large \@author\par}%
  \end{center}%
  \vskip 12pt
}
\makeatother
\newcommand{\supplementheading}{%
  \twocolumn[{%
    \titleformat{name=\section,numberless}{\centering\normalfont\fontsize{16}{19}\selectfont\bfseries}{}{0pt}{}%
    \section*{Supplemental Analyses}%
    \vspace{2pt}\hrule height .5pt\vspace{12pt}%
  }]%
}

\makeatletter
\newcommand{\makepaperfrontmatter}[1]{%
  \twocolumn[{%
    \@maketitle
    \begin{minipage}{\textwidth}%
      \normalfont\normalsize
      \hrule height .4pt
      \vspace{7pt}%
      \noindent\textbf{Abstract}---#1\par
      \vspace{7pt}%
      \hrule height .4pt
    \end{minipage}%
    \vspace{16pt}%
  }]%
}
\makeatother

\usepackage[hidelinks]{hyperref}
\usepackage{bookmark}
\usepackage{flushend}
\newcommand{\Elites}{\mathcal{E}}
\newcommand{\Candidates}{\mathcal{C}}
\title{EliteTable: Self-Improving Program Search\\for Executable Control Policies}
\author{%
  \textbf{Stuart Farmer}\\[3pt]
  {\normalsize Casa Dilauro Independent Research
  (\href{https://casadilauro.com}{casadilauro.com})}\\
  {\normalsize\href{mailto:stuart@casadilauro.com}{\texttt{stuart@casadilauro.com}}}\\[3pt]
  {\normalsize\href{https://github.com/StuartFarmer/elitetable}{\texttt{github.com/StuartFarmer/elitetable}}}%
}

\date{}
\begin{document}
\raggedbottom
\raggedend
\makepaperfrontmatter{%
EliteTable is a self-improving program-search algorithm guided by large language models (LLMs) that uses a fixed-capacity table of high-scoring solutions to guide new proposals, edits of existing solutions, and remixes of several solutions.
Performance was evaluated across six Gymnasium control tasks, each with ten generations and fifty candidate slots per generation. Each task was evaluated across ten independently initialized searches.
Candidate-score medians rose between the first and final generations in all six tasks, and selected-winner search-score medians rose in five. In \texttt{CartPole-v1}, the selected winner reached the search-score ceiling in the first generation of all ten searches.
All ten paired held-out endpoints improved in four of the six tasks. \texttt{BipedalWalker-v3} and \texttt{CarRacing-v3} improved without reaching their reporting targets. Six of ten \texttt{Acrobot-v1} searches declined on held-out episodes, and the median paired change was negative.
The results document improvement during program search and the transfer of selected-winner gains to new episodes in some tasks. Comparisons and ablations are needed to establish EliteTable's relative performance and the contributions of its components.
}

\section{Introduction}\label{introduction}\label{related-work}
This work was motivated by the broader study of recursive self-improvement and LLM-guided program search: how these algorithms are structured, which tasks they address, how well their solutions generalize, and which mechanisms support improvement.

The study examines one part of that motivation: whether a program-search algorithm inspired by prior work can also show improvement through repeated evaluation and revision.

In program search, one part of the improvement loop proposes a solution, often source code or an executable policy. Another part, the evaluator, provides a score or feedback on how that solution performs on the task.

\textit{Optimization by PROmpting (OPRO)} illustrates this feedback pattern: a language model receives a task description and previously evaluated solutions, then uses that feedback to propose further solutions~\cite{yang2024opro}.

\textit{FunSearch} makes this feedback loop concrete for executable programs. It uses an island-based database of solutions, maintaining multiple populations to support diversity. Its applications include cap-set constructions and online bin-packing heuristics~\cite{romeraparedes2024funsearch}.

\textit{AlphaEvolve} extends evolutionary code modification to scientific and computational problems, using evaluator feedback to guide revisions~\cite{novikov2025alphaevolve}. In these systems, evaluation gives program search an external measure of whether a proposed change is useful.

These algorithms differ in what they retain and how they generate candidates. \textit{ShinkaEvolve} combines parent sampling, code-novelty rejection, and bandit-based selection among language models~\cite{lange2025shinkaevolve}.

\textit{Evolution of Heuristics} evolves descriptions of heuristic ideas together with their executable implementations, giving the language model both conceptual and code-level information about the solutions it explores~\cite{liu2024eoh}.

EliteTable draws on these approaches through a fixed-capacity table of high-scoring programs. New candidates compete with retained solutions after each generation, while three operators generate fresh proposals, edits, and remixes. The design aims to provide a simple feedback loop for repeated program improvement.

Keeping incumbents eligible guarantees that the best recorded search score cannot decrease, but does not guarantee that newly generated solutions improve. A search may retain elites based on similar heuristics and become confined to a local optimum~\cite{romeraparedes2024funsearch}. Selected solutions may also fail to generalize to new episodes, so held-out evaluation is necessary~\cite{cawley2010selection}.

The study asks two questions. First, do in-sample winner scores and candidate-score distributions improve across generations? Second, do selected-winner gains transfer to held-out episodes? I examine these questions across six Gymnasium control tasks~\cite{towers2024gymnasium}.

Records were kept for every candidate across the sixty searches. Between generations one and ten, pooled candidate-score medians increased in all six tasks and median selected-winner search scores increased in five. \texttt{CartPole-v1} winners were already at the ceiling in generation one.

In \texttt{CarRacing-v3}, the candidate population showed a bimodal score distribution: a higher-return concentration moved toward stronger scores across generations while a low-return group persisted.

Held-out evaluation addressed the second question. All ten paired endpoint changes were positive in four tasks. \texttt{Acrobot-v1}'s median paired held-out change was negative. \texttt{BipedalWalker-v3} and \texttt{CarRacing-v3} improved between the first and final evaluations, but neither reached its reporting target.

Held-out gains were not necessarily monotonic: intermediate performance sometimes declined. Longer searches might produce further gains, but the tested budget does not establish whether \texttt{BipedalWalker-v3} or \texttt{CarRacing-v3} would eventually reach their targets.

\section{EliteTable Program Search}\label{elitetable-method}
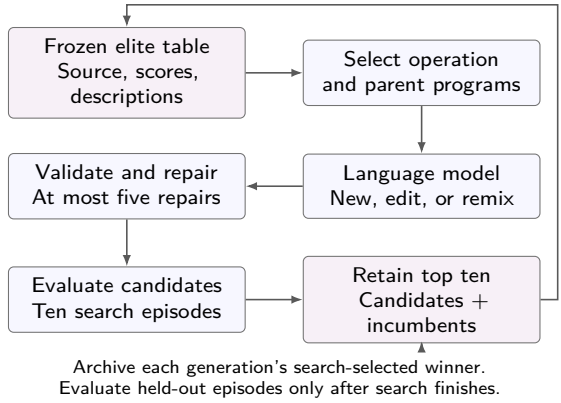
\begin{figure}[t]
\centering
\begin{tikzpicture}[
 box/.style={draw=black!55,rounded corners=2pt,fill=blue!3,
 font=\sffamily\footnotesize,align=center,text width=2.85cm,
 minimum height=0.85cm,inner sep=4pt},
 arr/.style={-{Latex[length=1.5mm]},draw=black!65,line width=.6pt}]
\node[box,fill=violet!6] (elite) at (0,0) {Frozen elite table\\Source, scores, descriptions};
\node[box] (select) at (3.9,0) {Select operation\\and parent programs};
\node[box] (llm) at (3.9,-1.5) {Language model\\New, edit, or remix};
\node[box] (valid) at (0,-1.5) {Validate and repair\\At most five repairs};
\node[box] (eval) at (0,-3) {Evaluate candidates\\Ten search episodes};
\node[box,fill=violet!6] (retain) at (3.9,-3) {Retain top ten\\Candidates + incumbents};
\draw[arr] (elite) -- (select);
\draw[arr] (select) -- (llm);
\draw[arr] (llm) -- (valid);
\draw[arr] (valid) -- (eval);
\draw[arr] (eval) -- (retain);
\draw[arr] (retain.east) -- (5.7,-3) -- (5.7,0.9) -- (0,0.9) -- (elite.north);
\node[font=\sffamily\scriptsize,align=center,text width=7.5cm] at (2,-4.05)
 {Archive each generation's search-selected winner.\\Evaluate held-out episodes only after search finishes.};
\draw[arr,dashed] (retain.south) -- (3.9,-3.55);
\end{tikzpicture}
\caption{EliteTable's evaluation--revision loop. All candidate slots finish
before the table is updated. Held-out scores do not enter the loop.}
\label{fig:elitetable-optimizer-loop}
\end{figure}

\textit{Figure~\ref{fig:elitetable-optimizer-loop}} shows the search loop, summarized
in \textit{Algorithm~\ref{alg:elitetable}}. The retained table supplies context for
subsequent proposals and preserves previously discovered high-scoring programs.

\begin{table*}[!t]
\centering\small
\setlength{\tabcolsep}{4pt}
\renewcommand{\arraystretch}{1.1}
\begin{tabularx}{\textwidth}{@{}l>{\raggedright\arraybackslash}p{0.21\textwidth}rr>{\raggedright\arraybackslash}X@{}}
\toprule
Environment & Observation/action interface & Step cap & Target & Task Settings \\
\midrule
\texttt{CartPole-v1} & State vector; discrete & 500 & 475 & Recorded environment profile.\\
\texttt{MountainCar-v0} & State vector; discrete & 200 & $-110$ & Recorded environment profile.\\
\texttt{Acrobot-v1} & State vector; discrete & 500 & $-100$ & Recorded environment profile.\\
\texttt{LunarLander-v3} & State vector; discrete & 1000 & 200 & Gravity $-10$; wind disabled.\\
\texttt{BipedalWalker-v3} & State vector; continuous & 1600 & 300 & Standard, non-hardcore task.\\
\texttt{CarRacing-v3} & $96\times96$ RGB; continuous & 1000 & 900 & Continuous control; lap-completion fraction 0.95; domain randomization disabled. Policies derive control directly from native RGB observations, so this task is interpreted separately from state-vector tasks.\\
\bottomrule
\end{tabularx}
\caption{Task suite, settings and descriptive reporting targets. Other constructor
settings follow the recorded environment profiles. RGB denotes red, green,
and blue image channels.}
\label{tab:environments}
\end{table*}

\subsection{Generation and repair}
EliteTable uses three operators: new, edit, and remix. Generation one contains only fresh proposals; the top $K$ evaluated solutions enter the elite table. Subsequent generations use the allocation in \textit{Table~\ref{tab:study-design}}. Fresh proposals receive elite descriptions and scores, edits receive one parent's source, and remixes combine information from up to three parents. Parents are sampled uniformly from the table frozen at the start of the generation.

Before evaluation, proposed solutions are checked for schema compliance, valid source edits, and parseable Python syntax using the abstract syntax tree (AST).

Solutions must satisfy the required policy interface and pass duplicate checks. Duplicate detection compares AST structures rather than source text.

Solutions are then passed to the evaluator. Validation or execution failures, such as access to an undefined variable, return diagnostics to a repair loop so the model can revise the proposal.

If the revised solution still fails, it returns to the repair loop, up to five repairs per candidate slot.

Candidate timeouts are treated as failures, not zero-reward episodes or successful truncations. Infrastructure failures interrupt execution rather than enter the candidate-repair loop.

\begin{algorithm}[t]
\caption{EliteTable search}\label{alg:elitetable}
\small
\begin{algorithmic}[1]
\Require Task, fixed LLM, search panel; $T=10$, $N=50$, $K=10$
\State $\Elites_0\gets\emptyset$
\For{$t=1,\ldots,T$}
\State Freeze parent table at $\Elites_{t-1}$
\State Allocate new/edit/remix slots (\textit{Table~\ref{tab:study-design}})
\State Apply empty/single-elite fallbacks; shuffle slots
\State Sample parents uniformly; remix parents are distinct
\State $\Candidates_t\gets\emptyset$
\ForAll{candidate slots}
\State Generate program; validate, evaluate, and repair
\State Add successful evaluated program to $\Candidates_t$
\EndFor
\State Wait for every slot to finish or be discarded
\State $\Elites_t\gets\operatorname{TopK}(\Elites_{t-1}\cup\Candidates_t)$
\State Archive the search-selected winner in $\Elites_t$
\EndFor
\State Evaluate archived winners on held-out episodes
\end{algorithmic}
\end{algorithm}

\subsection{Elite selection}
As briefly mentioned, EliteTable retains the highest-scoring programs across completed generations.
For elite capacity $K$, the update rule is

\begin{equation}
    \Elites_t =
    \operatorname{TopK}\left(\Elites_{t-1}\cup\Candidates_t\right).
\end{equation}

Here $\Elites_t$ is the retained elite set after generation $t$,
$\Candidates_t$ contains valid new candidates from that generation, and
$\operatorname{TopK}$ selects up to $K$ programs. Candidates are ranked by descending mean search reward.

For equal scores, edits take priority over remixes, followed by incumbent elites and then fresh proposals. Any remaining ties are resolved by ascending organism identifier.

This persistent elitist mechanism ensures that the highest recorded
\emph{search-panel} reward cannot decrease between completed generations,
assuming previously recorded evaluations remain unchanged. This monotonicity
does not imply monotonic held-out performance.

\section{Experimental Design}\label{experimental-design}
\subsection{Study structure}

Six control environments were selected from the Gymnasium reinforcement learning library~\cite{towers2024gymnasium}. Their task profiles are listed in \textit{Table~\ref{tab:environments}}.

The tasks were selected for their perceived differences in difficulty and their variety of observation and action spaces.

Across the sixty searches, hyperparameters were chosen to balance computational cost against the opportunity for improvement.

\begin{center}
\begin{minipage}{\linewidth}
\small\setlength{\tabcolsep}{3pt}
\renewcommand{\arraystretch}{1.1}
\captionsetup{hypcap=false}
\begin{tabular}{@{}p{0.47\linewidth}p{\dimexpr0.53\linewidth-6pt\relax}@{}}
\toprule
\textbf{Hyperparameter} & \textbf{Setting} \\
\midrule
Tasks & 6 \\
Searches per task & 10 \\
Population size & 50 \\
Elite table size & 10 \\
Generations & 10 \\
Initial operator split & 100\% new, 0\% edit, 0\% remix \\
Subsequent operator split & 20\% new, 40\% edit, 40\% remix \\
Search initialization seeds & $\{0,\ldots,9\}$ \\
In-sample seeds & $\{0,\ldots,9\}$ \\
Out-of-sample seeds & $\{1000,\ldots,1099\}$ \\
Maximum repairs per slot & 5 \\
Early stopping? & False \\
\bottomrule
\end{tabular}
\par\smallskip
\begin{minipage}{\linewidth}\footnotesize
The subsequent operator split applies when at least two elites are available.
Held-out episodes are evaluated per distinct generation winner within each search.
\end{minipage}
\captionof{table}{Shared search and evaluation settings.}
\label{tab:study-design}
\end{minipage}
\end{center}

Each candidate slot includes its initial proposal and any repair attempts; it either produces an evaluated policy or is discarded. Slot counts therefore differ from counts of valid programs, model requests, and completed episode evaluations.

The study records the configuration and execution history of each search. Initialization seeds control operator scheduling and parent sampling, but do not make hosted model responses deterministic. Reproduction can therefore follow the recorded procedure without guaranteeing identical search trajectories.

\subsection{Environments and policy interface}
Generated solutions must implement \texttt{Solution(Policy)}, a Python class with an asynchronous action method. Policy objects receive the current observation, observation and action spaces, and task instructions.

The policy interacts with the environment through an observe--act loop. Its runtime interface does not expose reward values or environment internals.

A policy may retain state within an episode, provided it satisfies the required interface. That state must be reset at the start of each episode.

Generation and repair prompts include task documentation, resolved environment settings, the observation/action contract, the objective, and execution constraints. This documented task knowledge is distinct from the information exposed to a policy during execution.

For \texttt{BipedalWalker-v3}, the prompts also include explicit observation indexing, normalization, motor ordering, and reward details. Prompt construction is configurable by task.

The study therefore evaluates program search with documented task knowledge, rather than learning solely from interaction.

\subsection{Model configuration}
I used the requested model route \path{openai/gpt-oss-120b:nitro} through OpenRouter for proposal generation and repair.

Responses were capped at 16,384 tokens per request.

The runner does not explicitly set temperature or top-$p$, so the study does not specify fixed values for those parameters. Returned model, provider, and usage metadata are retained when available. The requested route does not guarantee identical provider execution across requests.

Because fresh proposals receive elite summaries, they cannot serve as a history-independent sampling baseline.

\subsection{Fitness and held-out evaluation}
For policy $\pi$, its episode return under seed $s$ is the undiscounted sum

\begin{equation}
R(\pi,s)=\sum_{t=0}^{T_s-1}r_t,
\end{equation}

where $T_s$ is the number of steps before environment termination or the
configured step cap.

Candidate fitness is the arithmetic mean over the ten fixed search seeds
$S=\{0,\ldots,9\}$:

\begin{equation}
F(\pi)=\frac{1}{10}\sum_{s\in S}R(\pi,s).
\end{equation}

EliteTable ranks candidates by higher fitness scores. Tasks formulated as minimization problems would require an appropriate score transformation.

All candidate slots must finish evaluation or be discarded before the elite table is updated. The top $K$ programs are then selected from the new evaluated candidates and existing elites, using the tie-breaking order described above.

After search finishes, the fixed historical winner from every generation is
evaluated on 100 held-out episode seeds,
\[
H=\{1000,\ldots,1099\},
\]
which are disjoint from both the search panel and the previously inspected
pilot panel 100--199.

Each distinct selected policy is evaluated once per held-out seed within a search. Repeated generation winners within that search reuse cached measurements. Held-out outcomes do not influence generation, repair, promotion, stopping, or winner selection.

For environment $e$, search $i$, and generation $g$, held-out performance is

\begin{equation}
Y_{e,i,g}
=
\frac{1}{100}\sum_{s\in H}R(\pi^{*}_{e,i,g},s),
\end{equation}

where $\pi^{*}_{e,i,g}$ is the winner selected by search fitness.

A valid held-out mean requires finite returns on all 100 specified seeds and no recorded evaluation failures. Partial means are excluded. Missing evaluations remain missing; they are not replaced with zero or rerun.

The paired held-out change is $\Delta_{e,i}=Y_{e,i,10}-Y_{e,i,1}$. I report the median of these ten within-search changes for each task, rather than the difference between the first- and final-generation medians.

Descriptive 95\% percentile intervals use 10,000 bootstrap resamples of the ten whole-search paired records, with replacement. Each resample preserves the within-search pairing and recomputes the median change. These intervals describe variation conditional on the fixed episode panels and tested configuration; they do not establish robustness across new panels or models.

As a retrospective exploratory analysis, I applied two-sided exact paired sign tests~\cite{nistSignTest} to the ten endpoint changes per task. The null is $\Pr(\Delta>0\mid\Delta\ne0)=1/2$, not equality of mean returns. Exact zero changes are excluded from sign counts. Holm adjustment~\cite{holm1979} covers all six tasks, retaining an all-tied task as $p=1$ for correction only. These tests assume independent searches and condition on the fixed episode panels and tested configuration. They do not establish comparative superiority or causal operator effects.

\begin{figure*}[!t]
\centering
\includegraphics[width=\textwidth]{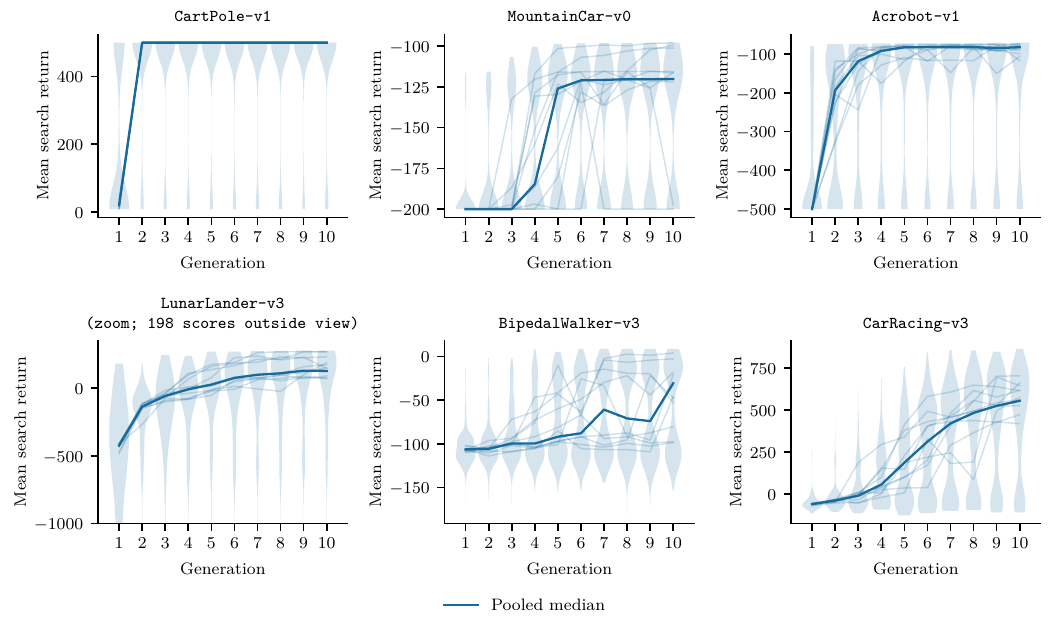}\caption{Candidate search-return distributions across generations, pooled over ten searches per task. Solid blue lines show pooled candidate medians; faint lines show medians within individual searches. Violin widths show smoothed density, not operator membership or independent replication.}
\label{fig:population-distributions}
\end{figure*}

\subsection{Execution and reproducibility}
Sixty searches were completed: ten initialization seeds across six tasks. Each began at generation zero with no existing solutions. Evaluation ran on an AMD64 host using eight worker processes, with up to 25 concurrent model operations for generation and repair. The configured deadlines were 45 seconds per episode, ten seconds per policy action, and 120 seconds per model-operation attempt.

The episode and action deadlines exclude Docker startup, queueing, and final cleanup. Generated policies ran in Docker containers with networking disabled, a read-only root filesystem, and resource limits.

Core reproduction code and recorded evidence are available at \url{https://github.com/StuartFarmer/elitetable}. The accompanying reproduction guide identifies the pinned revision and offline reconstruction commands. The retrospective sign-test extension and additional inputs and scripts for supplemental analyses accompany this manuscript locally; they are not included in that pinned public release. Reconstructing recorded results and running new stochastic searches are distinct procedures.

\begin{figure*}[!t]
\centering
\includegraphics[width=\textwidth]{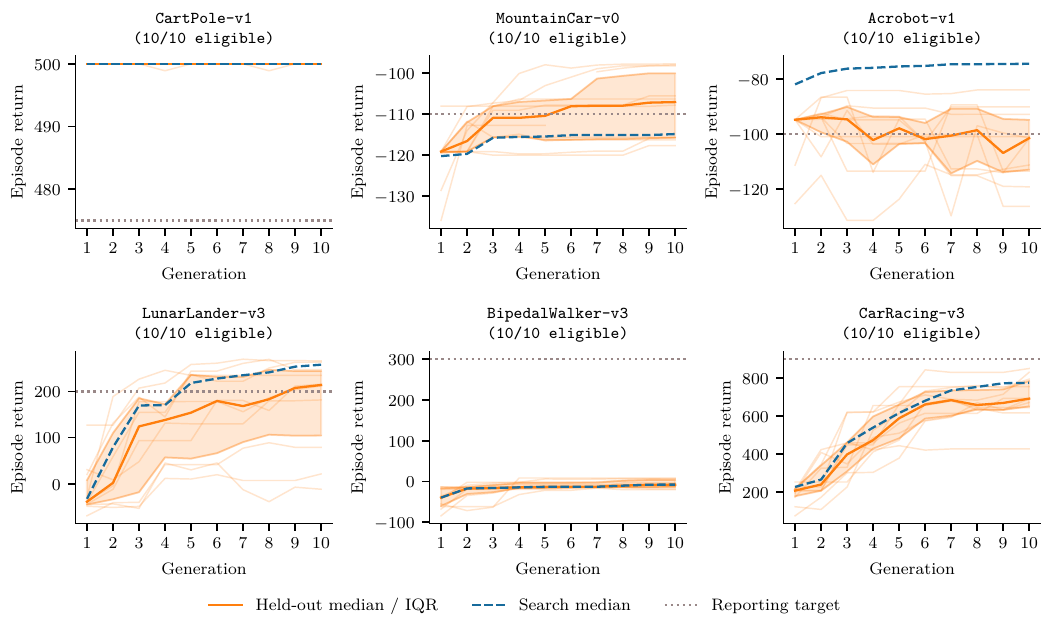}
\caption{Search and held-out returns over ten generations.
Dashed blue lines show median search return; solid orange lines show median
held-out return, with interquartile ranges (IQRs) shaded and individual search
trajectories drawn faintly. Dotted lines indicate the reporting threshold.
Each environment contains ten eligible searches.
\texttt{MountainCar-v0} generation 6 has nine valid held-out measurements; all other
environment-generation combinations have ten. Shading represents
between-search dispersion rather than a confidence interval.}
\label{fig:learning-curves}
\end{figure*}

\begin{table*}[!t]
\centering
\small
\setlength{\tabcolsep}{3pt}
\begin{tabular}{l r r r r r r}
\toprule
Environment & First & Final & Change [95\% CI] & Improved & Target & $p_{\mathrm{Holm}}$ \\
\midrule
\texttt{CartPole-v1}
  & 500.00 & 500.00
  & $0.00\;[0.00,\,0.00]$ & 0/10 & 10/10 & --- \\
\texttt{MountainCar-v0}
  & $-119.18$ & $-107.07$
  & $+7.41\;[2.56,\,22.47]$ & 10/10 & 6/10 & 0.01172 \\
\texttt{Acrobot-v1}
  & $-94.83$ & $-101.53$
  & $-6.38\;[-18.64,\,7.98]$ & 4/10 & 3/10 & 1.00000 \\
\texttt{LunarLander-v3}
  & $-39.04$ & 214.04
  & $+189.31\;[109.25,\,260.20]$ & 10/10 & 6/10 & 0.01172 \\
\texttt{BipedalWalker-v3}
  & $-39.30$ & $-7.65$
  & $+42.27\;[5.89,\,55.94]$ & 10/10 & 0/10 & 0.01172 \\
\texttt{CarRacing-v3}
  & 206.66 & 691.81
  & $+507.11\;[421.20,\,629.13]$ & 10/10 & 0/10 & 0.01172 \\
\bottomrule
\end{tabular}
\caption{Held-out performance across ten searches per task. First and Final are medians of the generation-one and generation-ten mean returns. Change is the median paired change with a descriptive 95\% bootstrap confidence interval (CI). Improved counts positive paired changes; Target counts final winners meeting the reporting threshold. Exploratory two-sided sign-test $p$-values are Holm-adjusted across all six tasks. A dash denotes ten ties and no informative test.}
\label{tab:heldout-results}
\end{table*}

\section{Experimental Results}\label{experimental-results}
\subsection{Completion and improvement during search}
All sixty searches completed. Held-out measurements were available for 599 of 600 generation-winner panels. The missing \texttt{MountainCar-v0} search-seed-8 panel was at generation six and did not affect any endpoint pair.

Median selected-winner search scores increased between the first and final generations in five environments. \texttt{CartPole-v1} winners were already at the score ceiling in generation one.

Newly evaluated candidate medians rose between generations one and ten in all six tasks, including \texttt{CartPole-v1}. \textit{Figure~\ref{fig:population-distributions}} shows these population changes, extending the evidence beyond retention of strong initial winners.

These distributions contain newly evaluated candidates, not retained elites or held-out winners. Each score averages ten search episodes; failed or discarded candidates without scores are excluded. The \texttt{LunarLander-v3} display is restricted to returns between $-1000$ and $350$: 198 scores outside this range remain in the calculations.

Candidate-score distributions show bimodality in some tasks. This motivated an exploratory analysis of whether fresh proposals, edits, and remixes contribute differently to the score mixture. The operator analyses appear in Supplemental Analyses.

\subsection{Performance on held-out episodes}
\textit{Table~\ref{tab:heldout-results}} reports paired endpoints and target counts. All ten searches improved in each of four environments: \texttt{MountainCar-v0}, \texttt{LunarLander-v3}, \texttt{BipedalWalker-v3}, and \texttt{CarRacing-v3}. \texttt{CartPole-v1} remained at the ceiling, while six of ten \texttt{Acrobot-v1} searches declined. \texttt{BipedalWalker-v3} and \texttt{CarRacing-v3} improved without a final winner reaching the reporting target.

Each of the four tasks with ten positive changes has raw $p=0.00195$ and Holm-adjusted $p=0.01172$. For \texttt{Acrobot-v1}, raw $p=0.75391$ and adjusted $p=1$, so the observed negative median does not establish a systematic decline. The ten \texttt{CartPole-v1} ties provide no informative sign test and do not establish equivalence.

\textit{Figure~\ref{fig:learning-curves}} shows how outcomes develop across generations. Held-out scores sometimes decline even when the endpoint improves. \texttt{Acrobot-v1}'s positive search gains coexist with a negative median paired held-out change, consistent with possible selection-panel specialization. The study shows transfer of endpoint gains to new episodes in four tasks, but does not establish robustness across task families or rule out overfitting. Broader evaluation and comparisons with other search and reinforcement-learning methods are needed.

\subsection{Candidate reliability and resource use}
Of 30,000 candidate slots, 29,959 reached evaluated status (99.9\%) and 41 were discarded. The logs contain 38,289 model calls, including 8,287 repair calls, or 21.6\% of all calls. Repair calls amount to 27.6\% of the 30,000 initial proposal calls; because a slot may undergo multiple repairs, this is not the percentage of proposals needing repair. The completion rate describes the combined procedure and does not isolate the repair loop's contribution.

\section{Discussion and Conclusion}\label{implications-and-discussion}
The rising candidate-score medians show progress beyond retaining a single strong initial solution. Newly generated populations also shift toward higher returns. In four tasks, every final selected winner improved over its first-generation counterpart on the fixed held-out episode panel. These endpoint gains do not by themselves rule out overfitting to the selection panel.

\texttt{Acrobot-v1} exposes a limitation: search scores rose while the median paired held-out change was negative. Further examination is needed to identify the cause. Held-out evaluation covers each generation's selected winner, not every retained elite, so it does not describe the full elite population. Search used ten episode seeds, compared with 100 held-out seeds. Future studies could test whether broader selection panels reduce specialization; the current design does not isolate that effect.

\texttt{BipedalWalker-v3} and \texttt{CarRacing-v3} improved without reaching their reporting targets. The tested budget does not establish whether additional generations would eventually solve these tasks. Their gains should therefore be distinguished from task attainment.

\subsection{Program search and neural-policy training}
EliteTable revises policy source code rather than updating neural-policy weights through policy gradients. Source edits can change numerical constants as well as program structure. Policy-gradient methods instead update parameterized policies, often neural networks, using estimates of the gradient of expected return with respect to policy parameters~\cite{sutton1999policygradient}.

The selected tasks are drawn from Gymnasium's standard reinforcement-learning environments~\cite{towers2024gymnasium}. Matched comparisons are needed to assess the relative strengths of program search and neural-policy training on these tasks.

Source revisions can alter control structures, numerical constants, state, and execution flow. Programmatic reinforcement learning can also combine program search with guidance from a trained neural policy~\cite{verma2018pirl}. These are different update mechanisms; the present study provides no matched comparison of their learning efficiency.

\subsection{Limitations and next steps}

Understanding EliteTable's components requires ablations and comparisons with other systems. The observed improvements cannot yet be attributed specifically to reuse, editing, remixing, retention, or repair. Ten searches per environment quantify variation under fixed episode panels, but do not establish robustness across episode distributions, language models, prompts, or search configurations. The tasks are well known, so the model may have encountered descriptions or potential solutions during training.

Time-based deadlines can produce different outcomes on different hardware. Hosted-provider variation can also change model outputs and search trajectories, even when the same model route is requested.

The final solutions are Python programs available for inspection, but source availability alone does not establish human interpretability.

I would prioritize budget-matched comparisons with other program-search algorithms, including a history-independent generation baseline, followed by complete-search ablations and comparisons across language models. Longer budgets, new episode panels, and other task families would test the limits of the observations. The present results provide development evidence for those designs rather than confirmatory evidence of comparative superiority.

\flushcolsend
\subsection{Conclusion}\label{conclusion}
EliteTable contributes to the study of self-improving LLM-assisted program search. Across sixty searches, candidate populations improved on selection episodes, while held-out evaluation distinguished transferable gains, saturation, and regression. This work supports further investigation of repeated program improvement as part of the broader agenda toward recursive self-improvement. It does not test a system that modifies its own optimizer or language model.

\clearpage
\raggedcolsend
\begingroup
\captionsetup{font=small,labelfont=it,skip=6pt,hypcap=false}
\setlength{\emergencystretch}{2em}
\setlength{\parskip}{0pt}
\raggedbottom
\setcounter{section}{0}
\setcounter{table}{0}
\setcounter{figure}{0}
\renewcommand{\thesection}{S\arabic{section}}
\renewcommand{\thetable}{S\arabic{table}}
\renewcommand{\thefigure}{S\arabic{figure}}
\renewcommand{\theHsection}{supplement.\arabic{section}}
\renewcommand{\theHtable}{supplement.\arabic{table}}
\renewcommand{\theHfigure}{supplement.\arabic{figure}}
\supplementheading
\noindent All analyses use the same sixty completed searches. Core reproduction code is available in the public GitHub repository. Prompts, policy sources, and additional analysis inputs accompany this manuscript locally; the reproduction guide identifies additions outside the pinned public release.

\section{Candidate populations during search}
\subsection{Population and winner endpoints}
I first compared candidate and selected-winner scores in the first and final generations to distinguish progress in newly generated programs from retention of strong incumbents. For each task, I pooled evaluated candidates across ten searches and calculated their median search score. Separately, I calculated the median score of the ten selected generation winners. Scores average returns on ten fixed search episodes; discarded candidates without scores are excluded.

\begin{center}
\small\setlength{\tabcolsep}{3pt}
\begin{tabular}{lrrrr}
\toprule
& \multicolumn{2}{c}{Candidates} & \multicolumn{2}{c}{Winners} \\
Environment & Gen. 1 & Gen. 10 & Gen. 1 & Gen. 10 \\
\midrule
\texttt{CartPole-v1} & 21.00 & 500.00 & 500.00 & 500.00 \\
\texttt{MountainCar-v0} & -200.00 & -120.25 & -120.30 & -114.90 \\
\texttt{Acrobot-v1} & -500.00 & -82.70 & -82.00 & -74.55 \\
\texttt{LunarLander-v3} & -423.92 & 125.80 & -32.79 & 258.04 \\
\texttt{BipedalWalker-v3} & -106.43 & -30.77 & -39.60 & -7.58 \\
\texttt{CarRacing-v3} & -60.03 & 555.07 & 226.05 & 775.30 \\
\bottomrule
\end{tabular}
\captionof{table}{Candidate and selected-winner medians at generations one and ten.}
\label{tab:search-progress}
\end{center}

New-candidate medians increased in all six environments. Selected-winner medians increased in five, as \texttt{CartPole-v1} winners were already at the ceiling. These results show progress in the programs being generated beyond retention of a strong early winner.

Generation one contains only fresh proposals, whereas later generations mix fresh proposals, edits, and remixes. The population changes include this composition change. This analysis does not isolate an operator effect or establish held-out improvement of the full candidate population.

\newpage
\subsection{\texttt{CarRacing-v3} distribution across generations}
I also examined the full \texttt{CarRacing-v3} candidate-score distribution to determine whether rising medians reflected a broad shift or the disappearance of weak candidates. For each generation, I pooled evaluated scores across ten searches and estimated a Gaussian kernel density with a common bandwidth of 25 return units. Multiplying each density by its evaluated-candidate count gives a band whose area equals that count.

The ten generations are stacked from bottom to top. A colored band's \emph{thickness} gives that generation's count density; its cumulative upper boundary is not the density. All 4,992 evaluated scores are included, with 498--500 candidates per generation. Eight discarded slots have no score. Scores average ten selection episodes, and smoothing extends beyond the observed score range.

\begin{center}
\includegraphics[width=\linewidth]{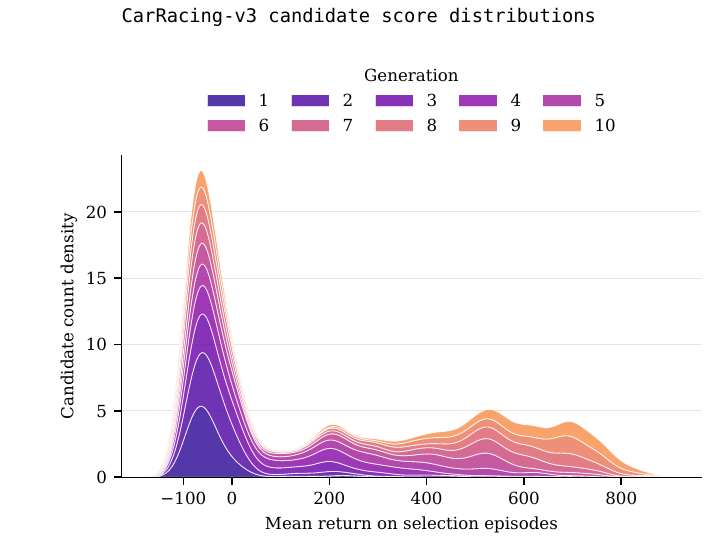}
\captionof{figure}{\texttt{CarRacing-v3} candidate distributions over ten generations.}
\label{fig:carracing-generation-distributions}
\end{center}

Higher-return concentrations emerge and move toward stronger scores as generations advance. Weak candidates persist, but the distribution shifts toward higher returns. These are in-sample changes under a changing operator mixture, not evidence of population-wide held-out gains.

\newpage
\section{Held-out outcomes and sensitivity}
\subsection{Paired search endpoints}
I evaluated each search's first and final selected winner on the same 100 held-out episodes. Winners were selected exclusively by search scores; held-out returns did not affect selection. Pairing within each search makes improvement and regression visible without pooling reward scales across tasks.

Hollow circles mark the first generation and filled points mark the final generation. Rightward movement indicates improvement; dashed lines mark reporting targets. Panel headings give the numbers of improved searches and final target-reaching winners. Each task retains its own reward scale.

\begin{center}
\includegraphics[width=\linewidth]{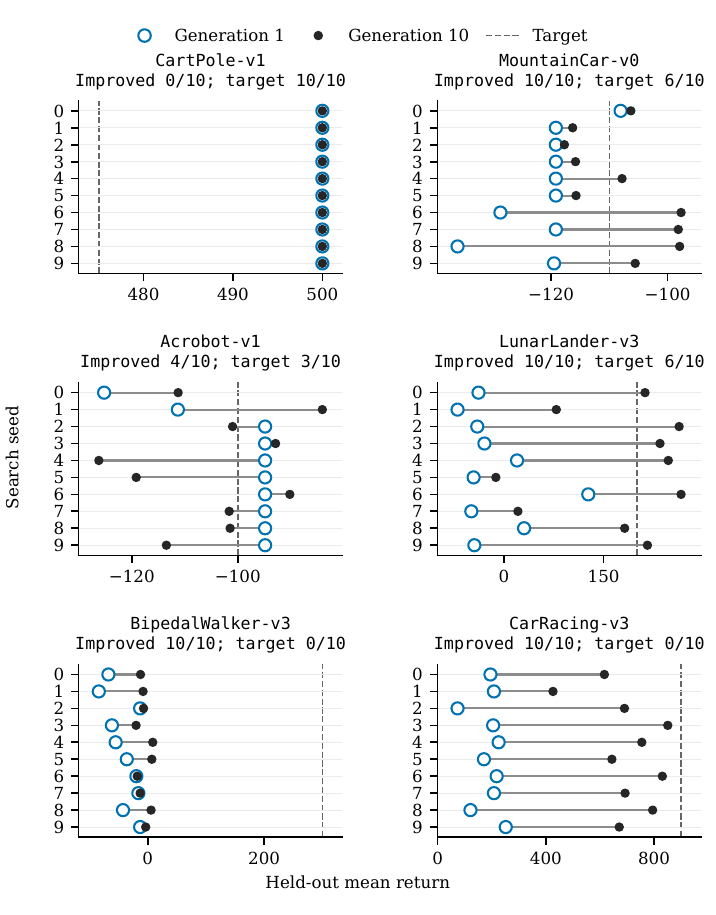}
\captionof{figure}{First and final held-out performance for all sixty searches.}
\label{fig:heldout-endpoints}
\end{center}

All ten endpoint pairs improved in \texttt{MountainCar-v0}, \texttt{LunarLander-v3}, \texttt{BipedalWalker-v3}, and \texttt{CarRacing-v3}. \texttt{CartPole-v1} remained at the ceiling, while six of ten \texttt{Acrobot-v1} searches declined. \texttt{BipedalWalker-v3} and \texttt{CarRacing-v3} improved without a final winner reaching the reporting target.

All sixty endpoint pairs are available. The missing \texttt{MountainCar-v0} search-seed-8
panel at generation six affects only the intermediate trajectory.

\newpage
\subsection{Sensitivity to individual searches}
For each task, I computed the ten paired held-out changes from generation one to generation ten. I omitted one search at a time and recomputed the median of the remaining nine changes. The minimum and maximum of these ten medians measure sensitivity to an individual search; they are not confidence intervals.

CP, MC and AC denote \texttt{CartPole-v1}, \texttt{MountainCar-v0} and \texttt{Acrobot-v1}; LL, BW and CR denote \texttt{LunarLander-v3}, \texttt{BipedalWalker-v3} and \texttt{CarRacing-v3}. The final rows give the full-sample median and leave-one-search-out (LOO) range.

\begin{center}\footnotesize\setlength{\tabcolsep}{4pt}
\textbf{(a) Classic-control tasks}\par\smallskip
\begin{tabular}{lrrr}\toprule
Search seed & CP & MC & AC \\
\midrule
0 & 0.00 & 1.76 & 13.98 \\
1 & 0.00 & 2.88 & 27.28 \\
2 & 0.00 & 1.46 & -6.14 \\
3 & 0.00 & 3.37 & 1.98 \\
4 & 0.00 & 11.36 & -31.40 \\
5 & 0.00 & 3.46 & -24.34 \\
6 & 0.00 & 31.01 & 4.67 \\
7 & 0.00 & 21.01 & -6.78 \\
8 & 0.00 & 38.13 & -6.61 \\
9 & 0.00 & 13.94 & -18.64 \\
\midrule
All ten: median & 0.00 & 7.41 & -6.38 \\
LOO min. & 0.00 & 3.46 & -6.61 \\
LOO max. & 0.00 & 11.36 & -6.14 \\
\bottomrule\end{tabular}\par\medskip
\textbf{(b) Box2D tasks}\par\smallskip
\begin{tabular}{lrrr}\toprule
Search seed & LL & BW & CR \\
\midrule
0 & 250.34 & 55.18 & 421.20  \\
1 & 148.46 & 76.03 & 218.29  \\
2 & 303.52 & 5.89 & 617.04  \\
3 & 263.97 & 41.56 & 645.68  \\
4 & 227.37 & 63.71 & 529.35  \\
5 & 33.48 & 42.99 & 473.15  \\
6 & 139.49 & 1.45 & 612.59  \\
7 & 70.05 & 3.06 & 484.87  \\
8 & 151.24 & 48.18 & 673.49  \\
9 & 260.20 & 9.58 & 419.54  \\
\midrule
All ten: median & 189.31 & 42.27 & 507.11  \\
LOO min. & 151.24 & 41.56 & 484.87  \\
LOO max. & 227.37 & 42.99 & 529.35  \\
\bottomrule\end{tabular}\par\medskip
\captionof{table}{Paired held-out changes and leave-one-search-out medians.}
\label{tab:paired-sensitivity}
\end{center}

Omitting any one search leaves the direction of every task's median change unchanged. This check concerns variation among the ten searches under common episode panels. It does not establish robustness to new episode panels, prompts, or models.

Endpoint gains also need not be monotonic. At least one adjacent-generation
decline occurred in one \texttt{CartPole-v1}, five \texttt{MountainCar-v0}, ten \texttt{Acrobot-v1}, ten
\texttt{LunarLander-v3}, seven \texttt{BipedalWalker-v3} and nine \texttt{CarRacing-v3} searches. \texttt{LunarLander-v3}'s
median fell from 179.38 at generation six to 168.29 at generation seven before
reaching 214.04. \texttt{CarRacing-v3} fell from 683.81 at generation seven to 658.00 at
generation eight before reaching 691.81.

\newpage
\section{Resource accounting}
I aggregated model-call logs, candidate repair counters, discarded slots, and recorded attempt durations across ten searches per task. All sixty searches began at generation zero and finished in one recorded attempt; none resumed a previous search.

For reward-scale context, I also evaluated a random-action policy on the 100-episode held-out panel and report the median of its run-level mean returns. This is not a budget-matched program-search baseline. Calls, repairs, discarded slots (Disc.), and hours are summed across searches. Hours include reporting as well as search.

\begin{center}\footnotesize\setlength{\tabcolsep}{2pt}
\begin{tabular}{lrrrrr}
\toprule
Task & Random & Calls & Repairs & Disc. & Hours \\
\midrule
\texttt{CartPole-v1} & 21.87 & 7134 & 2134 & 17 & 4.56 \\
\texttt{MountainCar-v0} & -200.00 & 6157 & 1157 & 0 & 6.05 \\
\texttt{Acrobot-v1} & -500.00 & 6546 & 1546 & 9 & 4.60 \\
\texttt{LunarLander-v3} & -200.81 & 6269 & 1269 & 3 & 4.05 \\
\texttt{BipedalWalker-v3} & -101.02 & 6064 & 1064 & 4 & 7.87 \\
\texttt{CarRacing-v3} & -32.68 & 6119 & 1117 & 8 & 29.76 \\
\midrule
Total & --- & 38289 & 8287 & 41 & 56.88 \\
\bottomrule
\end{tabular}

\captionof{table}{Random-action returns and recorded resources by task.}
\label{tab:reference-resources}
\end{center}

The logs contain 38,289 model calls, including 8,287 repair calls. Repairs
account for approximately 21.6\% of logged calls, so candidate-slot counts
understate model usage. Of 30,000 slots, 29,959 (99.9\%) reached evaluated status and
41 were discarded.

Recorded attempt durations total 56.88 hours, including 29.76 hours for \texttt{CarRacing-v3}. These are neither CPU-hours nor elapsed wall time for the whole study. Two logged calls lack usage records, so complete token and monetary totals remain unknown.

\newpage
\section{Operator distributions and elite admission}
\subsection{\texttt{CarRacing-v3} candidate distributions}
I grouped \texttt{CarRacing-v3} candidates by generation operator to examine how fresh proposals, edits, and remixes contributed to the score mixture. Generations two, six, and ten represent early, middle, and final post-initialization search, pooling ten searches at each generation.

For each operator and generation, I estimated a Gaussian kernel density with bandwidth 25 return units and multiplied it by the evaluated-candidate count. Band thickness gives count density; band area gives that operator's candidate count. Shared axes and bandwidth support comparison. Counts appear above each panel; the discarded slot at generation six is excluded. Smoothing extends beyond observed returns.

\begin{center}
\includegraphics[width=\linewidth]{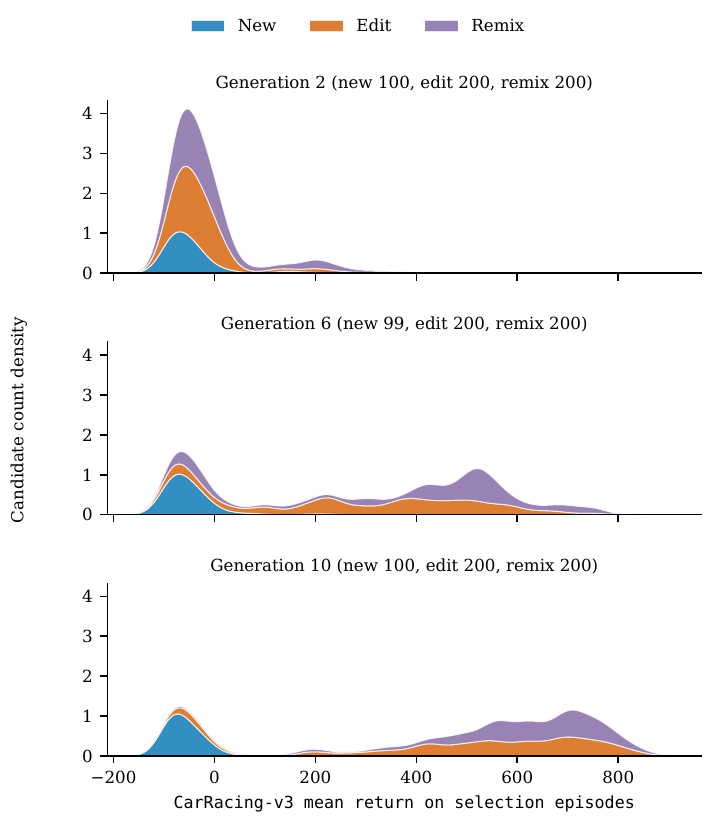}
\captionof{figure}{\texttt{CarRacing-v3} score distributions by generation operator.}
\label{fig:carracing-operators}
\end{center}

All three operators contribute to the early low-score concentration. By generation ten, fresh proposals remain concentrated at low scores while edits and remixes supply most of the higher-score population. Unequal allocation, parent selection, and inherited information prevent attributing these differences solely to operator choice.

\newpage
\subsection{Admission to the elite table}
I extended the analysis by measuring how often an evaluated candidate entered the elite table in its birth generation, pooling generations 2--10 across ten searches per task. I grouped admissions by the operator used to generate the candidate. Dividing admissions by each operator's evaluated-candidate count accounts for unequal allocations. This measures entry into the table, not the duration of later retention.

Points give admission percentages, with admitted and evaluated counts noted
alongside. Discarded slots are excluded from the rate denominators.

\begin{center}
\includegraphics[width=\linewidth]{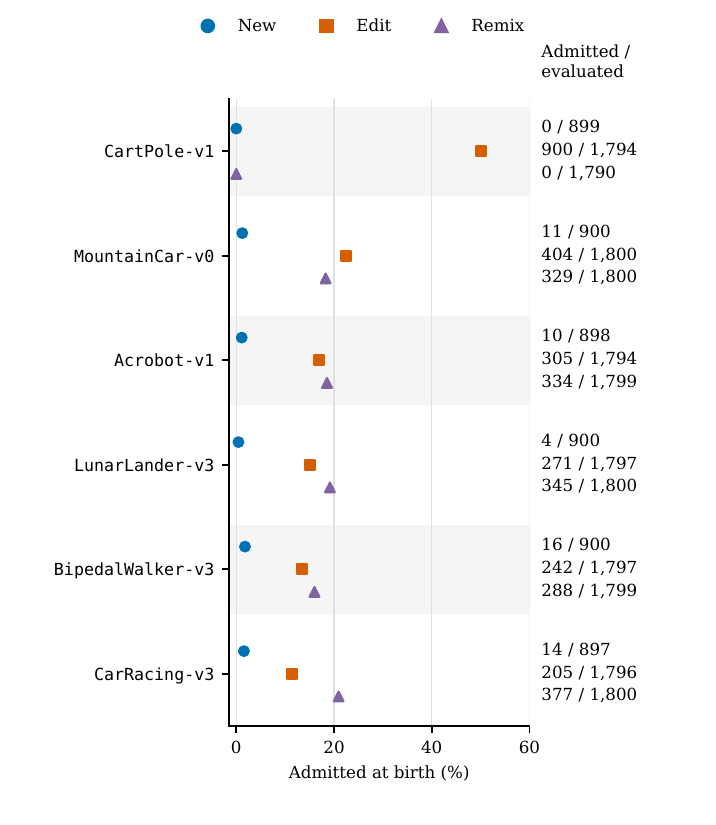}
\captionof{figure}{Birth-generation elite-admission rates by task and operator.}
\label{fig:operator-admission}
\end{center}

Only 55 of 5,394 evaluated fresh proposals entered the elite table, approximately 1\%. Fresh proposals had lower observed admission rates under the tested configuration. They receive elite summaries but no parent source, so they are not a history-independent sampling baseline.

Incumbent competition and deterministic tie-breaking also affect admission. \texttt{CartPole-v1}'s equal-score priority for edits is especially visible. The rates do not identify causal operator effects, and candidates are dependent observations within searches.

\newpage
\subsection{Complete operator counts and returns}
I also summarized candidates by operator within each task over generations 2--10. The table records slots, evaluated candidates (Eval.), discarded slots (Disc.), median search returns, and birth-generation elite admissions (Adm.). Each search return averages ten selection episodes. Score medians exclude discarded slots; slot totals include them.

\begin{center}\small\setlength{\tabcolsep}{3pt}
\begin{tabular}{lrrrrr}\toprule
\multicolumn{6}{l}{\textbf{\texttt{CartPole-v1}}} \\
Operator & Slots & Eval. & Disc. & Median & Adm. \\
fresh & 900 & 899 & 1 & 33.70 & 0  \\
edit & 1800 & 1794 & 6 & 500.00 & 900  \\
remix & 1800 & 1790 & 10 & 500.00 & 0  \\
\midrule
\multicolumn{6}{l}{\textbf{\texttt{MountainCar-v0}}} \\
Operator & Slots & Eval. & Disc. & Median & Adm. \\
fresh & 900 & 900 & 0 & -200.00 & 11  \\
edit & 1800 & 1800 & 0 & -121.80 & 404  \\
remix & 1800 & 1800 & 0 & -120.30 & 329  \\
\midrule
\multicolumn{6}{l}{\textbf{\texttt{Acrobot-v1}}} \\
Operator & Slots & Eval. & Disc. & Median & Adm. \\
fresh & 900 & 898 & 2 & -500.00 & 10  \\
edit & 1800 & 1794 & 6 & -84.70 & 305  \\
remix & 1800 & 1799 & 1 & -81.50 & 334  \\
\midrule
\multicolumn{6}{l}{\textbf{\texttt{LunarLander-v3}}} \\
Operator & Slots & Eval. & Disc. & Median & Adm. \\
fresh & 900 & 900 & 0 & -361.13 & 4  \\
edit & 1800 & 1797 & 3 & 68.25 & 271  \\
remix & 1800 & 1800 & 0 & 80.70 & 345  \\
\midrule
\multicolumn{6}{l}{\textbf{\texttt{BipedalWalker-v3}}} \\
Operator & Slots & Eval. & Disc. & Median & Adm. \\
fresh & 900 & 900 & 0 & -109.16 & 16  \\
edit & 1800 & 1797 & 3 & -77.02 & 242  \\
remix & 1800 & 1799 & 1 & -64.61 & 288  \\
\midrule
\multicolumn{6}{l}{\textbf{\texttt{CarRacing-v3}}} \\
Operator & Slots & Eval. & Disc. & Median & Adm. \\
fresh & 900 & 897 & 3 & -66.11 & 14  \\
edit & 1800 & 1796 & 4 & 270.51 & 205  \\
remix & 1800 & 1800 & 0 & 391.17 & 377  \\
\bottomrule\end{tabular}

\captionof{table}{Operator counts, median search returns and elite admissions.}
\label{tab:operator-summary}
\end{center}

Fresh proposals had lower median returns than both edits and remixes in every environment. Admissions count elite membership at the candidate's birth generation, not later retention. These are descriptive associations: incumbent competition, parent selection, and tie-breaking prevent attributing the differences solely to operator choice.

\endgroup

\end{document}